\documentclass{article}

\usepackage[dblblindworkshop,final]{neurips_2026}
\workshoptitle{VLM4RWD}

\setcitestyle{numbers,square,comma}
 
\usepackage[utf8]{inputenc} % allow utf-8 input
\usepackage[T1]{fontenc}    % use 8-bit T1 fonts
\usepackage[hidelinks]{hyperref}      % hyperlinks
\usepackage{url}            % simple URL typesetting
\usepackage{booktabs}       % professional-quality tables
\usepackage{array}          % column specifiers for the prompt table
\usepackage{amsfonts}       % blackboard math symbols
\usepackage{amsmath}        % math environments
\usepackage{nicefrac}       % compact symbols for 1/2, etc.
\usepackage{microtype}      % microtypography
\usepackage{xcolor}         % colors
\usepackage{graphicx}       % figures
\usepackage{pgfplots}       % ablation graphs
\pgfplotsset{compat=1.18}
\usepackage{caption}
\usepackage{placeins}
\PassOptionsToPackage{hyphens}{url}% \FloatBarrier: keep floats inside their own section
 
\title{A Unified Evaluation Framework of Physical Adversarial Robustness in the Edge VLM Era}
 
\author{%
  Eric Liang$^{1}$ \quad
  Alan Luo$^{2,3}$\thanks{Corresponding author.} \quad
  Kaiwen Yuan$^{4}$ \\
  $^{1}$Saratoga High School \\
  $^{2}$Google \\
  $^{3}$University of Illinois Urbana-Champaign \\
  $^{4}$Mecka.ai \\
  \texttt{\{liae8407@lgsstudent.org, alanluo3@illinois.edu, kaiwen@mecka.ai\}}
}
 
\begin{document}
 
\maketitle
 
\begin{abstract}
Recent advancements in Vision-Language Models (VLMs) enable end-to-end visual grounding via language prompts and generative decoding, demonstrating largely improved zero-shot and open-vocabulary detection capabilities. As a result, VLMs are rapidly being deployed at the edge in safety-critical real-world systems. However, the physical adversarial robustness of edge VLMs has not been systematically studied, and the fundamental differences in output with modern object detectors create methodological challenges for fair evaluation. To compare the two families directly, we evaluate every selected representative model on one outcome: whether an adversarial patch worn on a person evades detection by the model. We report this by introducing a unified metric, conditional attack success rate (cASR), counted only on images the model still localizes under a matched gray occluder, and partition every failure into omission (no box), substitution (a box on the patch), and displacement (a box on neither). We benchmark the physical adversarial vulnerability of compact grounding VLMs against a range of object detectors, and find that \textbf{the VLMs can be as vulnerable as the detectors}. We disclose that \textbf{failure mode is largely associated with the output format of a model rather than its architecture family}. Of the three, only omission is visible to a downstream system; displacement and substitution return confident, well-formed boxes, making them potentially more dangerous failures. Further, we find that, under distribution shift, VLMs that primarily failed by omission instead shift to failing more by displacement/substitution. Thus, a grounding VLM can inherit the detector's vulnerability while showing the possibility of adopting more dangerous failure modes. Our code and data are available at \url{https://github.com/ericliang1/physical-adversarial-robustness}.
\end{abstract}
 
\section{Introduction}
 
Vision-language models (VLMs) are transitioning from describing a scene to localizing within one. Given a natural-language reference to an object, a VLM returns a bounding box. In embodied and autonomous settings~\citep{brohan2023rt, shridhar2022cliport}, a planner then conditions on that box directly~\citep{deruyttere2019talk2car}, with no confidence score to threshold and no signal of uncertainty, so a correct box and an incorrect one arrive in the same form. Recent grounding VLMs are increasingly built for real-world deployment, with compact sub-4B variants suited to embedded accelerators~\citep{zhu2025internvl3, steiner2024paligemma, yu2026minicpm, bai2025qwen3, liquidai2026lfm25vl}. Existing edge inference pipelines, however, were built around object detectors, and whether a grounding VLM is equally robust in their place cannot be settled on clean images, where both families localize a person at high accuracy. We therefore induce failures by training a physical adversarial patch against each model, a printed texture designed to suppress detection when worn on a person in the real world---an established threat to detectors that has not been tested against grounding VLMs on a common scale~\citep{thys2019fooling, brown2017adversarial}. The suppression objective of each adversarial patch drives down its corresponding model's native output, making adversarial robustness directly comparable across architectures and exposing the inherent vulnerability of each model's spatial grounding.

\begin{figure}[ht]
\centering
\includegraphics[width=\linewidth]{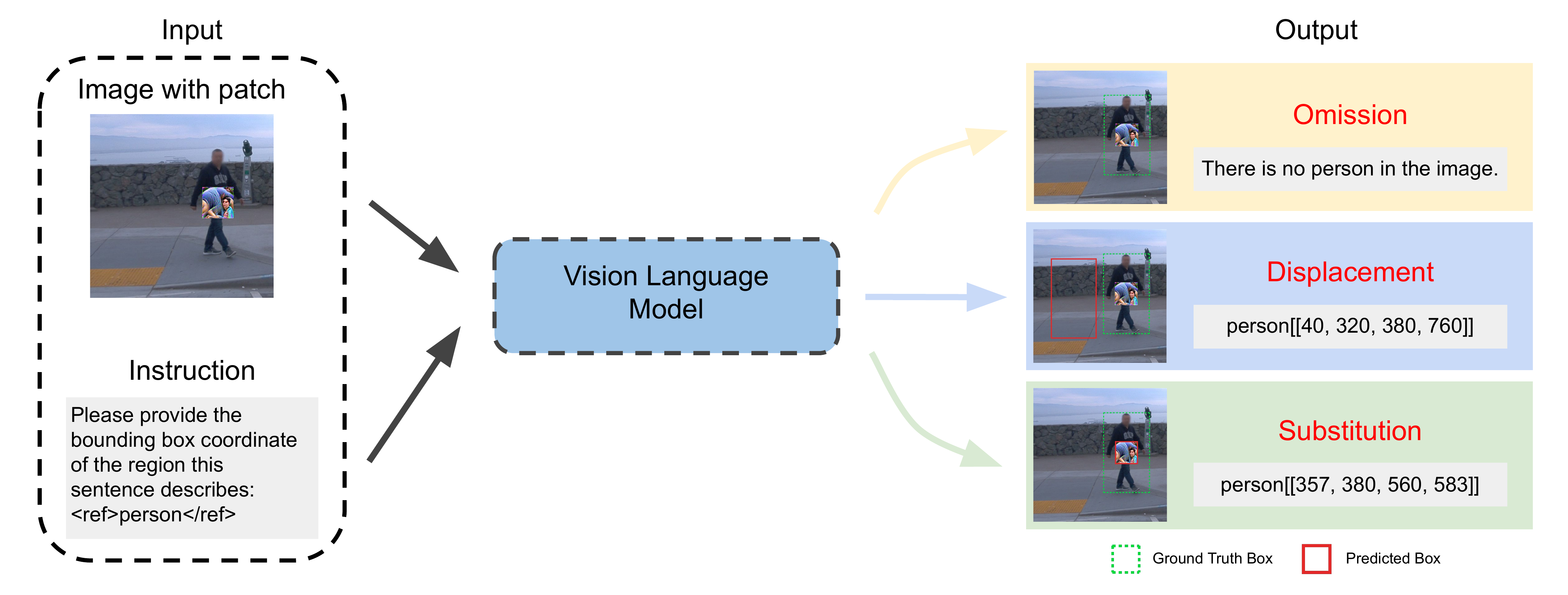}
\caption{The three failure modes: omission, displacement, and substitution.}
\label{fig:headline}
\end{figure}

We compare the two families across eleven models (Table~\ref{tab:main}) on the one predicate they share, whether a box lands on the target. We measure failure by introducing conditional attack success rate (cASR), which scores a patch only on images the model localizes under a matched gray occluder, and partition grounding failure into three modes---\textbf{omission} (no box), \textbf{substitution} (a box on the patch), and \textbf{displacement} (a box on neither); Figure~\ref{fig:headline} illustrates all three. We apply partitions in an occlusion-controlled comparison of the two families across held-out and out-of-distribution images.
 
Two findings follow. First, grounding VLMs can be as vulnerable as the detectors under adversarial patches in deployment settings. Substituting one for the other does not necessarily reduce a system's failure rate under a white-box attack. Second, the two families do not fail in the same way. Whether an attacked model omits or displaces is instead largely associated with its output format rather than its architecture family, while architecture also plays a role in the failure partition.
 
% Unlike omission, a displaced box appears as a well-formed detection to any downstream system---an important distinction for real-world deployment.
 
\section{Related work}
 
\subsection{Object detection and grounding VLMs}
 
 Traditional object detectors \citep{girshick2015fast, redmon2016you} have centered on convolutional neural networks (CNNs) for their superior capability for edge-device deployment due to low inference latency and excellent energy efficiency. The recent success of the transformer \citep{vaswani2017attention} has inspired vision transformer (ViT) based vision tasks \citep{dosovitskiy2020image, he2022masked} that rethink the architecture in a natural language processing schema. Shortly after, transformer-based object detectors \citep{carion2020end} came to dominate leaderboards~\citep{zhang2022dino, zong2023detrs}, demonstrating accuracy and generalization on large datasets. On the other hand, transformers still suffer from a lack of inductive biases and weaker performance on small and low-data problems \citep{11298561, dosovitskiy2020image}. Therefore, both CNN and transformer models are commonly used for edge deployment. To make detectors more general purpose, open-vocabulary capabilities have been integrated into CNN ~\citep{cheng2024yolo} and transformer architectures~\citep{liu2024grounding}, allowing detections to be driven by natural language prompts.

More recently, general purpose large language models have trended towards multimodality and have thus also emerged into visual grounding capabilities \citep{zhu2025internvl3, steiner2024paligemma, yu2026minicpm, team2026kimi, bai2025qwen3}. These grounding models not only yield bounding box coordinates of objects but also query objects in an open-vocabulary manner. Even though most open-weight multimodal large language models (MLLMs) and VLMs are very heavy, with billions, and in the largest cases trillions, of parameters \citep{team2026kimi}, there are also a handful of compact VLM variants that are promising for deployment in lightweight settings~\citep{yu2026minicpm, liquidai2026lfm25vl}. As a result, it may be a matter of time before edge devices are armed with VLMs either in place or in combination with detectors. Understanding the security of models in these two different detection schemas is therefore critical.
 
\subsection{Adversarial attacks on object detection}

The first adversarial attacks on modern object detectors date back to 2017, focusing on imperceptible perturbations applied to an image \citep{xie2017adversarial}. Later studies focused on transferability and generalization~\citep{thys2019fooling, brown2017adversarial} of adversarial attacks. These works utilized adversarial patches -- textures capable of existing in diverse scenes in the real world, revealing a model's vulnerabilities to specific localized patterns. The same idea has been further extended from the RGB domain to other (and mixed) modalities~\citep{abdelfattah2021towards, abdelfattah2021adversarial}. Underneath the variety, the threat model is consistent: the attacker optimizes pixels in texture space against a model's native output.

However, language models admit no such formulation. Gradients do not act on discrete tokens, so attacks there are found by searching over candidate strings, with gradients through the token embeddings guiding the search \citep{wallace2019universal, zou2023universal}. The form of an attack on a language interface is dramatically different: it can be typed, injected or retrieved.

VLMs sit at the intersection of both methods and are exposed to both vulnerabilities. Early work showed that CLIP-style encoders can classify objects by text in the image rather than by the object itself \citep{goh2021multimodal}, while later attacks introduced traditional imperceptible perturbation-based adversaries that transfer across models and prompts \citep{zhao2023attackvlm, luo2024cropa}. A few recent works have begun to target grounded localization directly through training-time backdoors, though they are digital rather than physical~\citep{raj2025trap, li2026iag}. On the other hand, vision-language-action models, which use VLM backbones to act in robotics, have been demonstrated to fail by physical attacks in their visual input~\citep{wang2025exploring, lu2026robots}; however, these attacks target the model's actions rather than its localization.

Despite these advances, understanding the physical vulnerabilities of a model's localization requires evaluating the way traditional detector attacks do~\citep{thys2019fooling}. Existing physical-attack evaluations cover detectors alone~\citep{lian2024padetbench}; none of these works addresses real-time VLMs small enough to be deployed on edge devices, which our work aims to address.

\section{Methodology}

\subsection{Models and datasets}
 
For models, representative traditional closed-set object detectors~\citep{khanam2024yolov11, ultralytics2026yolo26, lv2024rtdetrv2}, open-vocabulary detectors~\citep{cheng2024yolo, zhao2024open, liu2024grounding}, and edge VLMs~\citep{zhu2025internvl3, steiner2024paligemma, yu2026minicpm, bai2025qwen3, liquidai2026lfm25vl} whose parameter counts are no larger than 4B are selected for evaluation.
 
We conduct our training on three datasets for people detection---COCO~\citep{lin2014coco}, PASCAL VOC~\citep{everingham2010pascal} and AdvT-Shirt-1K~\citep{wang2024diffpatch}---and each contains ground-truth labels for every human in each scene. COCO and PASCAL VOC contain diverse person images in various scenes while AdvT-Shirt contains people with adversarial patches worn on shirts. However, we treat AdvT-Shirt the same as COCO and PASCAL VOC, as its garments were optimized against older detectors and have no effect on the models we test, as seen in the clean column of Table~\ref{tab:main}. We manually filter each dataset into subsets that contain only images with exactly one person. This ensures that every model is evaluated on a matched single-target detection task, removing ambiguity from VLM prompts, as VLMs generally output a single bounding box. Across all models and in-distribution datasets we use an 80/20 training/validation split, giving 1{,}324 training and 331 held-out validation images. We additionally evaluate out-of-distribution on 999 filtered single-person images from the Waymo Open Dataset~\citep{sun2020scalability}, which represent street-level pedestrian views captured from a moving vehicle. Exact sources, filter counts, and split construction are given in Appendix~\ref{app:data}. For our experiments, all images are resized to the native input resolution of the vision model: all VLMs use $448 \times 448$ while all detectors use $640 \times 640$.
 
\subsection{Adversarial patch training}
 
We train on 1{,}324 images---622 COCO, 539 PASCAL VOC, 163 AdvT-Shirt---each containing exactly one person. For each image, a single adversarial patch is applied to the center of the person. The patch is a square texture with an original resolution of $160 \times 160$. When placed, it is resized relative to the person ground-truth label according to $s_i = a \cdot \min(w_i, h_i)$, where $s_i$ is the side length of the placed patch in image $i$, $w_i$ and $h_i$ denote the width and height of the target bounding box, and $a$ denotes the scaling factor, with $a = 0.5$ for our main experiment. This scale allows the patch to cover the center of the person while leaving important person features such as heads and limbs fully visible, representing the size of a feasibly wearable garment.

\begin{figure}[t]
\centering
\includegraphics[width=\linewidth]{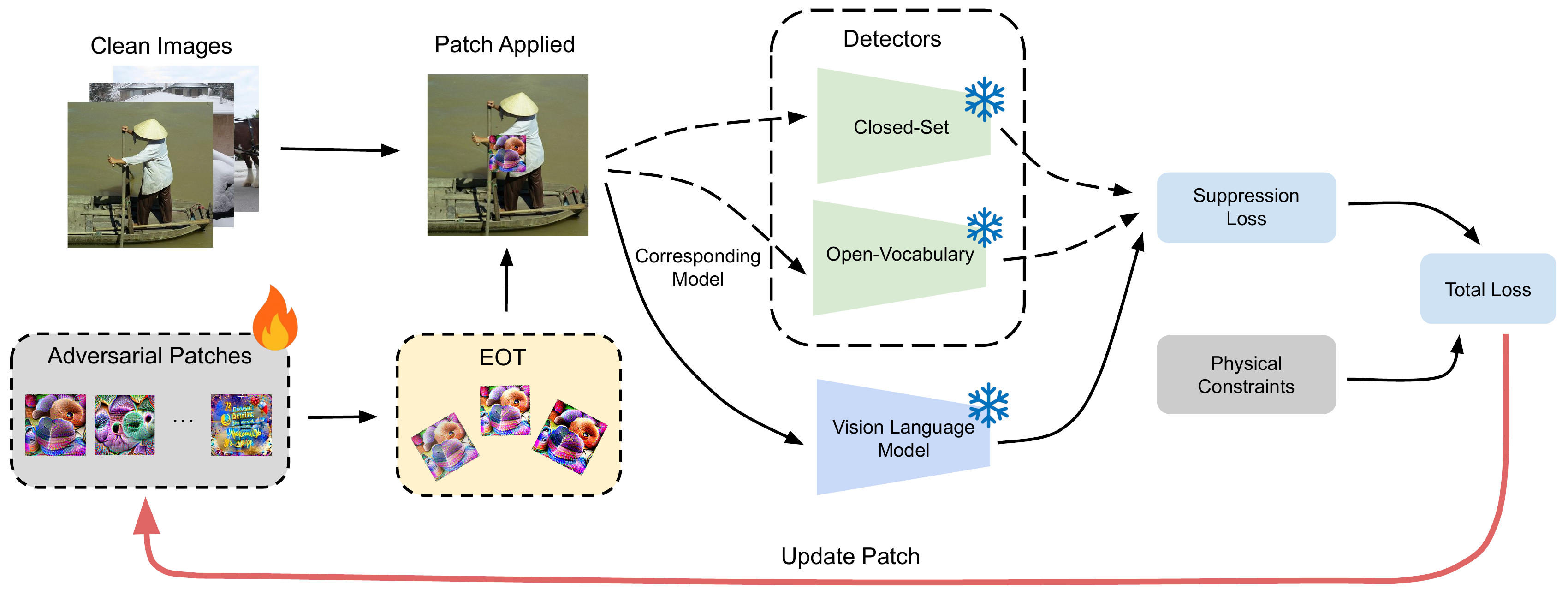}
\caption{Patch training pipeline.}
\label{fig:pipeline}
\end{figure}

\begin{figure}[t]
\centering
\includegraphics[width=1\linewidth]{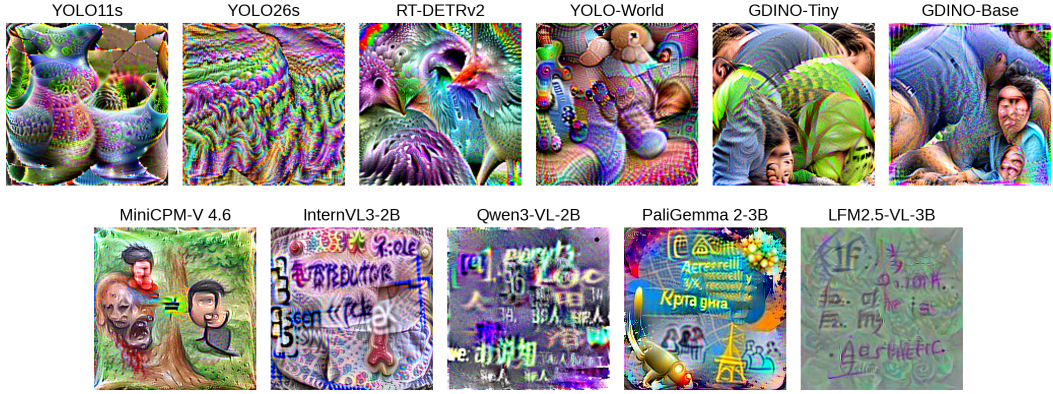}
\caption{Trained adversarial patches for each model. Patches trained on detectors contain high frequency patterns while patches trained on VLMs contain features that resemble text.}
\label{fig:patches}
\end{figure}

We use white-box optimization to independently train an image-agnostic patch for each model, as summarized in Figure~\ref{fig:pipeline}. Each attack is optimized to directly suppress the output of its respective target model. For detectors, patches are trained to suppress the final detection of the person. For VLMs, patches are trained to suppress the bounding box tokens, reducing the model's likelihood of emitting the spatial information of the patched individual. The resulting patches are shown in Figure~\ref{fig:patches}.
 
Optimization is performed across 500 epochs with a learning rate of 0.01 for all models, allowing all patches to converge and ensuring a fair comparison between them. To make the patches physically robust and effective across different scenes, we train with Expectation Over Transformation (EOT)~\citep{athalye2018synthesizing}, applying random rotations drawn uniformly from $[-20^\circ, 20^\circ]$ and multiplicative brightness jitter drawn uniformly from $[0.9, 1.1]$ (i.e., $\pm 10\%$). The patch scale is applied when the patch is placed on the target and is not part of the EOT distribution. All patches are applied digitally; ``physical'' represents the threat model rather than the evaluation medium.
 
The patch is trained using the loss function
\begin{equation}
L = L_s + \alpha L_{nps} + \beta L_{tv},
\end{equation}
where $L_s$ suppresses the output, $L_{nps}$ enforces printability~\citep{sharif2016accessorize}, $L_{tv}$ enforces smoothness, and $\alpha$ and $\beta$ are two hyperparameters to balance the scale. We set $\alpha = 0.01$ and $\beta = 0.01$, keeping the printability and smoothness terms small enough to constrain the patch to be physically realizable without trading away attack strength. Adapted from Thys et al.~\cite{thys2019fooling}, the non-printability and smoothness terms are
\begin{gather}
L_{nps} = \frac{1}{|P|}\sum_{p \in P}\ \min_{c \in C}\ (p - c)^2, \\
L_{tv} = \frac{1}{N}\sum_{i,j}\left(\lvert p_{i,j}-p_{i+1,j}\rvert + \lvert p_{i,j}-p_{i,j+1}\rvert\right),
\end{gather}
where $p_{i,j}$ denotes a patch pixel, $p$ a pixel of the patch $P$, $c$ a color in the set of printable colors $C$, and $|P|$ the number of patch pixels; in $L_{tv}$, each absolute difference is taken per color channel (red, green, blue) and $N$ is the total number of differences. For each model, $L_{tv}$ and $L_{nps}$ remain the same, and are used to constrain the patch to be physically realizable, while $L_s$ performs the same suppression task on the final differentiable prediction scores before the model's final output decision, which is unique to each model's architecture and native output structure. Concretely, $L_s$ takes one of three forms,
\begin{equation}
\label{eq:ls}
L_s =
\begin{cases}
\max_{k}\, z_k & \text{closed-set detector},\\[2pt]
\max_{k}\, \operatorname{sim}(r_k, t) & \text{open-vocabulary detector},\\[2pt]
\log p_\theta\!\left(b \mid q\right) & \text{vision--language model},
\end{cases}
\end{equation}
where $z_k$ is the person confidence of proposal $k$; $\operatorname{sim}(r_k, t)$ is the similarity between region $k$ and the ``person'' text embedding $t$; and $p_\theta(b \mid q)$ is the likelihood of decoding the target's box-opening token $b$ under prompt $q$. Minimizing $L_s$ suppresses the person confidence, the region--text similarity, or the grounding likelihood, respectively. 
 
\section{Evaluation framework}
 
We create a unified evaluation framework that aligns adversarial attack evaluation across heterogeneous model architectures. The framework ensures that the effects of adversarial attacks can be directly compared by framing success as the ability of a patched target individual to evade detection.
 
\subsection{Configuration}
 
The 331 validation images comprise 155 COCO, 135 PASCAL VOC, and 41 AdvT-Shirt; the 999 Waymo images are evaluated separately as the out-of-distribution set. Unless stated otherwise, the following default conditions hold: the overlap threshold is $\tau = 0.5$, the standard detection criterion~\citep{everingham2010pascal}; the patch is placed at the scale it was trained at; and detectors are read at their canonical confidence threshold.
 
\subsection{Output alignment}
 
To align with detectors, we prompt each VLM to produce one bounding box per image in its native grounding format, then normalize all outputs to a standard xyxy format. Any output that is malformed or fails to correspond to a valid bounding box is treated as a detection failure, enforcing evaluation based on spatial correctness. Each model is scored on any of its reported boxes $\hat{b}_i$: for a detector, it is every person box clearing $\gamma$; for a VLM, it is each box it emits.
 
Since traditional object detectors produce many detections per target instance ranked by confidence scores, while vision-language models often produce a single detection, a detector is read at a chosen confidence threshold whereas a VLM is read at its output directly. For VLMs, we decode greedily and do not consider the temperature, which is standard for single-target detection tasks. For detectors, we evaluate at the confidence thresholds at which their family is conventionally operated, ensuring that all models are evaluated in deployment-relevant settings. Closed-set detectors are read at the confidence threshold $\gamma = 0.5$. Open-vocabulary detectors are read at a lower confidence threshold, at $\gamma = 0.3$.
 
\subsection{Occlusion control and conditional ASR}
 
In evaluation, a gray digital patch can cause a failure by occluding part of the target, causing a detection to fail. As a result, standard attack success rate (ASR) can overestimate the adversarial effect of a patch, while subtraction-based flooring methods may underestimate the adversarial effect when the gray patch itself causes a high failure rate. To separate true adversarial success from occlusion and baseline model error, we evaluate each adversarial patch only on the images for which the corresponding model can successfully detect the person under the gray-patched condition. We define this metric as conditional ASR (cASR). Let $G$ be the set of images the model localizes correctly (Intersection over Union, IoU $\geq \tau$, against the ground-truth box $b_i$) under the gray occluder, and let $f_i = 1$ when $\mathrm{IoU}(\hat{b}_i, b_i) < \tau$, where $\hat{b}_i$ is the box predicted on the adversarially patched image. Then
\begin{equation}
\label{eq:casr}
\mathrm{cASR} = \frac{1}{|G|} \sum_{i \in G} f_i,
\end{equation}
the fraction of gray survivors that the adversarial patch breaks.
 
The floor's magnitude corresponds to localization tightness: tight boxes lose overlap once the target center is covered, whereas loose ones retain enough to clear $\tau$. The models most affected by the control are therefore the most precise rather than the weakest---and a benchmark reporting raw attack success would penalize them for that precision.
 
\subsection{Failure partition}
\label{sec:partition}
 
Conditional attack success quantifies failure frequency but not failure mode. Omission is tracked when no box of a proper format forms. Substitution is tracked when a box is formed and closely aligns with the patch, measured by $\mathrm{IoU}(\hat{b}_i, F_i) \geq 0.5$, where $F_i$ is the patch footprint. Displacement is tracked when a box is formed but lands on neither the target nor the patch, falling below $\tau$ against $b_i$ and below $0.5$ against $F_i$. The three are exclusive by construction and exhaustive by the last case, so they sum to cASR.
 
\section{Results}
 
\paragraph{Baseline localization.}
 
Table~\ref{tab:main} reports localization with no patch applied, where a model counts as correct if it emits a box overlapping the ground truth at IoU $\geq 0.5$. The criterion is identical for every model. In-distribution accuracy spans 83.0\% to 100.0\%. On the out-of-distribution Waymo set accuracy drops for every model, reflecting the harder detection task of street-level pedestrian views from a moving vehicle. Most models remain well above 65\%, while MiniCPM-V 4.6 falls to 51.1\% and InternVL3-2B to 27.6\% on the same images. By construction, cASR counts a failure only on images the model localizes correctly under the matched gray occluder (Eq.~\ref{eq:casr}), so no failure below is attributable to baseline localization. This conditioning means each model's Waymo cASR is read over its own, smaller gray-control set. For InternVL3-2B and MiniCPM-V 4.6, which localize driving pedestrians poorly, this set is as small as 305 and 366 of 999 images, respectively.
 
\begin{table}[t]
\centering
\footnotesize
\setlength{\tabcolsep}{4pt}
\caption{Clean detection accuracy and conditional attack success (cASR). Bold marks the lowest cASR within each family.}
\label{tab:main}
\begin{tabular}{@{}lrcccccccc@{}}
\toprule
 & & \multicolumn{6}{c}{In-distribution (trained)} & \multicolumn{2}{c}{Out-of-distribution} \\
\cmidrule(lr){3-8}\cmidrule(l){9-10}
 & & \multicolumn{2}{c}{COCO} & \multicolumn{2}{c}{VOC} & \multicolumn{2}{c}{AdvT-Shirt} & \multicolumn{2}{c}{Waymo} \\
\cmidrule(lr){3-4}\cmidrule(lr){5-6}\cmidrule(lr){7-8}\cmidrule(l){9-10}
Model & Params & Clean & cASR & Clean & cASR & Clean & cASR & Clean & cASR \\
\midrule
\multicolumn{10}{@{}l}{\textbf{Closed-set detectors}}\\
YOLO11s        & 9.5\,M   & 96.1  & 80.3          & 91.9 & 66.7          & 100.0 & 100.0         & 74.5 & 86.1          \\
YOLO26s        & 9.5\,M   & 95.5  & 66.1          & 92.6 & \textbf{50.5} & 100.0 & 100.0         & 75.5 & \textbf{38.7} \\
RT-DETRv2      & 20.2\,M  & 94.2  & \textbf{63.0} & 92.6 & 53.8          & 100.0 & \textbf{75.6} & 65.8 & 80.4          \\
\addlinespace
\multicolumn{10}{@{}l}{\textbf{Open-vocabulary detectors}}\\
YOLO-World     & 76.2\,M  & 96.1  & \textbf{25.6} & 97.8 & \textbf{15.0} & 100.0 & \textbf{2.5}  & 75.2 & \textbf{25.3} \\
GDINO-Tiny     & 172.3\,M & 97.4  & 64.2          & 97.8 & 56.1          & 100.0 & 85.4          & 89.8 & 62.1          \\
GDINO-Base     & 232.8\,M & 100.0 & 60.0          & 97.8 & 47.7          & 100.0 & 58.5          & 90.7 & 40.2          \\
\addlinespace
\multicolumn{10}{@{}l}{\textbf{Vision-language models}}\\
MiniCPM-V 4.6  & 1.3\,B   & 94.8  & \textbf{10.1} & 83.0 & \textbf{10.3} & 95.1  & \textbf{2.5}  & 51.1 & 9.6           \\
InternVL3-2B   & 2.1\,B   & 98.7  & 59.2          & 91.1 & 41.5          & 97.6  & 100.0         & 27.6 & 35.4          \\
Qwen3-VL-2B    & 2.1\,B   & 99.4  & 22.2          & 94.8 & 22.6          & 100.0 & 7.5           & 90.3 & 39.7          \\
PaliGemma 2-3B & 3.0\,B   & 97.4  & 59.3          & 94.1 & 37.6          & 100.0 & 97.6          & 86.7 & 15.4          \\
LFM2.5-VL-3B   & 3.1\,B   & 96.1  & 16.4          & 91.1 & 11.4          & 100.0 & 14.6          & 87.8 & \textbf{5.8}  \\
\bottomrule
\end{tabular}
\end{table}
 
\subsection{VLMs can be as vulnerable as detectors}
 
Table~\ref{tab:failure} reports cASR (Eq.~\ref{eq:casr}), computed over the gray-control set, so occlusion alone contributes near zero. Grounding VLMs rank among the most affected models in the benchmark: InternVL3-2B and PaliGemma 2-3B fail on 57.7\% and 55.2\% of the gray-control set, higher than two of the six detectors we test. AdvT-Shirt gives the cleanest comparison, as every model localizes at 95\% or above with no patch applied. With the adversarial patch applied, InternVL3-2B and PaliGemma 2-3B fail on 100\% and 97.6\% of their gray-control sets, comparable to the worst detectors (Table~\ref{tab:main}), YOLO11s and YOLO26s, which both reach 100\%. Substituting a VLM for a detector therefore does not necessarily lower the failure rate.
 
The most robust models also come from both families. MiniCPM-V 4.6 is the most robust overall, failing on only 9.2\% of the gray-control set. YOLO-World, though a detector, sits at 18.2\%, more robust than three of the five VLMs. The VLMs span 9.2\% to 57.7\%, while the detectors span 18.2\% to 78.3\%. The two ranges largely overlap, with the variation within each family exceeding the difference between them. Robustness is therefore largely a property of the individual model rather than its family.
 
The difference between the most and least affected models therefore reflects what the attack can suppress. The attack drives some quantity down (a detector's confidence, a VLM's token likelihood), which removes the box only when a single token gates the box's existence (Eq.~\ref{eq:ls}). Producing no box guarantees a failure, whereas perturbing coordinates only constitutes a failure if the box moves far enough to fall below $\tau$. InternVL3-2B and PaliGemma 2-3B open their output with such a gating token, so driving down its likelihood produces an omission; they are the two most affected VLMs we test. A free-text model has no such token. It writes coordinates as ordinary text that the decoder produces whether or not the target is visible, so the attack cannot delete the box, only mislocalize it. The free-text VLMs (LFM2.5-VL-3B, MiniCPM-V 4.6, and Qwen3-VL-2B) are also the three least affected as measured by cASR.
 
\subsection{Failure modes}
 
Table~\ref{tab:failure} partitions each model's failures into omission, substitution, and displacement (Section~\ref{sec:partition}).
 
\begin{table}[t]
\centering
\footnotesize
\setlength{\tabcolsep}{5pt}
\caption{Failure partition into omission (O), substitution (S), and displacement (D).}
\label{tab:failure}
\begin{tabular}{@{}llcccccccc@{}}
\toprule
 & & \multicolumn{4}{c}{In-distribution} & \multicolumn{4}{c}{Waymo (out-of-distribution)} \\
\cmidrule(lr){3-6}\cmidrule(l){7-10}
Model & Output Format & cASR & O & S & D & cASR & O & S & D \\
\midrule
\multicolumn{10}{@{}l}{\textit{Closed-set detectors}}\\
YOLO11s       & score-gated & 78.3 & 72.8 & 0.0  & 5.5  & 86.1 & 81.5 & 0.0  & 4.6  \\
YOLO26s       & score-gated & 65.3 & 62.2 & 0.0  & 3.1  & 38.7 & 31.0 & 0.0  & 7.6  \\
RT-DETRv2     & score-gated & 61.9 & 53.5 & 0.0  & 8.4  & 80.4 & 61.2 & 0.5  & 18.7 \\
\addlinespace
\multicolumn{10}{@{}l}{\textit{Open-vocabulary detectors}}\\
YOLO-World    & text-match  & 18.2 & 15.6 & 0.7  & 1.8  & 25.3 & 22.4 & 0.3  & 2.6  \\
GDINO-Tiny    & text-match  & 63.6 & 49.4 & 3.4  & 10.8 & 62.1 & 8.8  & 16.6 & 36.7 \\
GDINO-Base    & text-match  & 54.9 & 29.9 & 10.7 & 14.3 & 40.2 & 1.4  & 18.0 & 20.9 \\
\addlinespace
\multicolumn{10}{@{}l}{\textit{Vision-language models}}\\
MiniCPM-V 4.6 & free-text   & 9.2  & 0.7  & 0.3  & 8.1  & 9.6  & 0.0  & 0.3  & 9.3  \\
InternVL3-2B     & gated       & 57.7 & 56.7 & 0.0  & 1.0  & 35.4 & 4.6  & 1.3  & 29.5 \\
Qwen3-VL-2B      & free-text   & 20.1 & 0.0  & 1.5  & 18.7 & 39.7 & 0.0  & 0.9  & 38.8 \\
PaliGemma 2-3B    & gated       & 55.2 & 53.4 & 0.0  & 1.9  & 15.4 & 8.9  & 0.1  & 6.3  \\
LFM2.5-VL-3B  & free-text   & 14.3 & 0.0  & 5.2  & 9.1  & 5.8  & 0.0  & 1.1  & 4.7  \\
\bottomrule
\end{tabular}
\end{table}
 
\paragraph{Omission and displacement.} The two VLM output formats occupy opposite extremes of the failure mode partition. The gated VLMs, PaliGemma 2-3B and InternVL3-2B, fail almost entirely by omission: 96.7\% and 98.3\% of their failures produce no box, a higher share than any detector we test. The score-gated detectors behave the same way, with the YOLO models failing by omission on 93\% to 95\% of failures. At the other end, the free-text VLMs Qwen3-VL-2B and LFM2.5-VL-3B record exactly zero omissions at every threshold we sweep, and MiniCPM-V 4.6 fewer than 1\%. What these models produce instead is displacement, a box that parses correctly but lands on neither the person nor the patch. Notably, Qwen3-VL-2B and LFM2.5-VL-3B share an output format yet fail at different rates and split differently between displacement and substitution (displacement accounts for 18.7\% of Qwen3-VL-2B's 20.1\% and 9.1\% of LFM2.5-VL-3B's 14.3\%). Therefore, under a shared suppression objective that lowers the likelihood of the target box without targeting an empty response, output format is associated with whether a box is emitted, but not how often or where it lands. Alternative prompts can still shift this balance, as MiniCPM-V 4.6 omits on 7--9\% of attacked images under alternative phrasings (Table~\ref{tab:prompt}).
 
Because output format fixes emission but not placement, a stricter overlap criterion should affect only the models that emit a box. Raising the IoU threshold ($\tau$) from 0.5 to 0.9 (Figure~\ref{fig:iou} in Appendix~\ref{app:ablations}), the free-text models' failure rates climb sharply while the gated models' barely move. Qwen3-VL-2B's cASR reaches 95.6\%---the largest aggregated in-distribution cASR any model reaches in the benchmark---while its omission holds at exactly zero throughout, the entire increase being fully attributable to boxes it still emits. PaliGemma 2-3B and InternVL3-2B only move from 55.2\% to 68.6\% and 57.7\% to 62.8\%, remaining omission-dominated at every point. Thus raising $\tau$ does not affect omissions: with no box emitted, they fail at every threshold regardless. It instead converts displaced boxes into failures as they stop clearing the threshold.
 
\paragraph{Substitution.} A box that lands on the patch itself rather than the person similarly occurs in models whose emission is not gated on the target's presence. Every closed-set detector and both gated VLMs (PaliGemma 2-3B, InternVL3-2B) register exactly zero. Every other model registers a nonzero rate: the region-text matchers (GDINO-Base 10.7\%, GDINO-Tiny 3.4\%, YOLO-World 0.7\%) and the free-text VLMs (LFM2.5-VL-3B 5.2\%, Qwen3-VL-2B 1.5\%, MiniCPM-V 4.6 0.3\%). Substitution cuts across architectures and arises in the models where neither a confidence threshold nor a gating token withholds the box.
 
In deployment, substitution and displacement carry worse consequences than omission because both emit a box. A downstream system can detect an omission, since a model that returns nothing signals its own failure. Substitution and displacement, on the other hand, deliver a confident, well-formed box at an incorrect location, which may be indistinguishable from a correct detection. Substitution is arguably the more severe of the two. It inverts the attack, supplying a stronger signal that the model reports as the person rather than suppressing the person's own, so the system's attention is anchored to an object the attacker carries rather than to empty space. Its box also takes the patch's size, no larger than the person's at every scale we evaluate, so any spatial feature a system infers from the bounding box (such as distance) is systematically distorted. A single success rate from prior work collapses these distinctions, directly motivating our proposed failure partition.
 
\subsection{Under distribution shift}
 
The Waymo set holds every model to the same task on a domain none of the patches were trained against, approximating the distribution shift of deployment. cASR moves in both directions, so the distribution shift is not a uniform increase in difficulty. What changes is the failure mode. In-distribution, InternVL3-2B, GDINO-Base, and GDINO-Tiny fail predominantly by omission. On the Waymo set, their omission rate drops sharply (InternVL3-2B from 56.7\% to 4.6\%, GDINO-Base from 29.9\% to 1.4\%, and GDINO-Tiny from 49.4\% to 8.8\%) while their overall failure rates fall much less in comparison. Their remaining failures instead register as displacement or substitution, the modes a downstream system cannot detect. InternVL3-2B's displacement rises from 1.0\% to 29.5\%, GDINO-Tiny's substitution from 3.4\% to 16.6\% and its displacement from 10.8\% to 36.7\%, and GDINO-Base's substitution from 10.7\% to 18.0\%. Counterintuitively, InternVL3-2B's cASR decreases (57.7\% to 35.4\%), yet most of its remaining failures are now undetectable.
 
The closed-set detectors exhibit this shift far less: YOLO11s, YOLO26s, and RT-DETRv2 all remain omission-dominated on Waymo. In contrast, the open-vocabulary detectors and grounding VLMs do lose detectability. The one exception is YOLO-World, the most robust among them, which stays omission-dominated throughout.
 
Further, gated VLMs lose a key safety property under the distribution shift: in-distribution, InternVL3-2B and PaliGemma 2-3B fail by omission on 98.3\% and 96.7\% of their failures. On Waymo, PaliGemma 2-3B's omission falls from 53.4\% to 8.9\% and InternVL3-2B's nearly vanishes. Among the grounding VLMs, these two models are the only ones that predominantly fail visibly by omission, yet even that largely disappears under the distribution shift---highlighting that grounding VLMs can be as vulnerable as, or more vulnerable than, detectors in deployment settings.
 
\FloatBarrier
 
\section{Limitations}
 
Our adversarial attacks are digitally simulated and are not evaluated in physical settings. The patches are optimized to be printable and pose-robust, which is what ``physical'' refers to throughout; no printed patch or camera capture was evaluated, so we do not measure physical-world transfer. In addition, we only utilize white-box attack methods, which assume full access to the model and do not reflect real-world deployment scenarios. Future work should focus on evaluating the adversarial effect of physically printed patches and test black-box methods. Furthermore, we test only a suppression objective, and future work should evaluate persistence of the failure-mode partitions under different attack objectives.

\section{Conclusion}
 
Under one attack framework, one occlusion control, and a single shared predicate, \textbf{edge grounding VLMs can be as vulnerable as the classic detectors, yet they do not fail in the same way}: the form of each failure is largely associated with the model's output format rather than its architecture family. This difference carries the deployment consequence. A patch trained to suppress the model's detection of the person produces no box in some models and a well-formed, incorrectly placed box in others. Only the first is visible to a downstream system, while the second may be indistinguishable from a correct detection. Any grounding model bound for deployment should therefore be evaluated on two axes: how often it fails under physical attack, and whether those failures are visible in its output. The first requires an occlusion-controlled measurement, which our benchmark provides. Therefore, as evidenced by our controlled comparison, the shift from detectors to grounding VLMs does not reduce a system's vulnerability to physical attack. Under the distribution shift of real-world deployment, it may render the system even more vulnerable---failing in modes that are invisible to the downstream systems relying on it.
% ---------------------------------------------------------------------------
% REFERENCES. Uses natbib, which neurips_2026.sty loads by default.
% The Reference section does not count towards the page limit.
% ---------------------------------------------------------------------------
 
\begin{ack}
We thank our reviewers for helpful discussions. This work received no external funding.
\end{ack}
 
\bibliographystyle{plainnat}
\bibliography{reference}
 
% ===========================================================================
% APPENDIX (excluded from the nine-page main-paper limit)
% ===========================================================================
\newpage
\appendix
\FloatBarrier
 
\section{Ablation studies}
\label{app:ablations}
 
We report four ablations, each varying a single parameter of the attack while holding the rest of the pipeline constant: patch scale, overlap threshold, VLM prompt, and detector confidence threshold. All four are evaluated on the 331 validation images at the canonical settings, conditioned on the gray-control set.
 
\subsection{Patch size ablation}
 
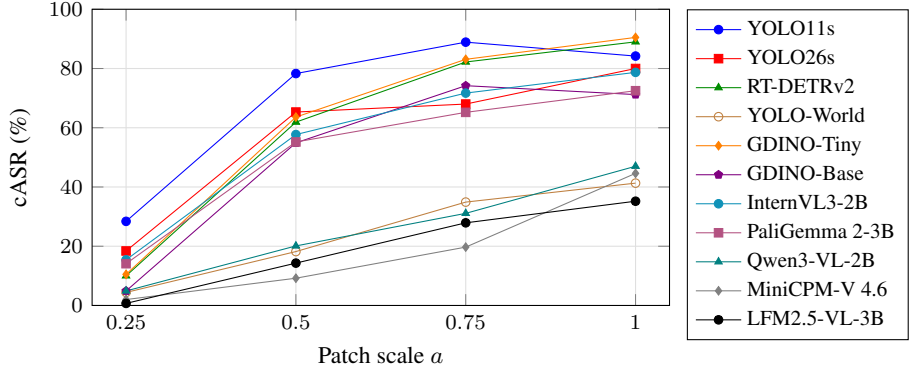
\begin{figure}[ht]
\centering
\begin{tikzpicture}
\begin{axis}[
    width=0.66\linewidth, height=5.5cm,
    xlabel={Patch scale $a$},
    ylabel={cASR (\%)},
    xtick={0.25,0.5,0.75,1.0},
    xmin=0.2, xmax=1.05,
    ymin=0, ymax=100,
    legend style={at={(1.03,1)}, anchor=north west, font=\footnotesize},
    legend cell align={left},
    grid=major, grid style={gray!25},
    mark size=1.6pt,
    tick label style={font=\footnotesize},
    label style={font=\small},
]
\addplot[blue, mark=*]                 coordinates {(0.25,28.4)(0.5,78.3)(0.75,88.9)(1.0,84.2)}; \addlegendentry{YOLO11s}
\addplot[red, mark=square*]            coordinates {(0.25,18.4)(0.5,65.3)(0.75,68.0)(1.0,80.0)}; \addlegendentry{YOLO26s}
\addplot[green!55!black, mark=triangle*] coordinates {(0.25,10.0)(0.5,61.9)(0.75,82.2)(1.0,89.0)}; \addlegendentry{RT-DETRv2}
\addplot[brown, mark=o]                coordinates {(0.25,4.4)(0.5,18.2)(0.75,34.9)(1.0,41.3)}; \addlegendentry{YOLO-World}
\addplot[orange, mark=diamond*]        coordinates {(0.25,10.5)(0.5,63.6)(0.75,83.1)(1.0,90.5)}; \addlegendentry{GDINO-Tiny}
\addplot[violet, mark=pentagon*]       coordinates {(0.25,4.9)(0.5,54.9)(0.75,74.2)(1.0,71.2)}; \addlegendentry{GDINO-Base}
% VLM patch-size sweeps (all models on the pooled COCO+VOC+AdvT-Shirt gray-survivor set)
\addplot[cyan!70!black, mark=*]          coordinates {(0.25,15.4)(0.5,57.7)(0.75,71.7)(1.0,78.7)}; \addlegendentry{InternVL3-2B}
\addplot[magenta!70!black, mark=square*] coordinates {(0.25,14.1)(0.5,55.2)(0.75,65.2)(1.0,72.5)}; \addlegendentry{PaliGemma 2-3B}
\addplot[teal, mark=triangle*]           coordinates {(0.25,4.9)(0.5,20.1)(0.75,31.1)(1.0,47.0)}; \addlegendentry{Qwen3-VL-2B}
\addplot[gray, mark=diamond*]            coordinates {(0.25,2.0)(0.5,9.2)(0.75,19.7)(1.0,44.6)}; \addlegendentry{MiniCPM-V 4.6}
\addplot[black, mark=otimes*]            coordinates {(0.25,0.7)(0.5,14.3)(0.75,27.9)(1.0,35.2)}; \addlegendentry{LFM2.5-VL-3B}
\end{axis}
\end{tikzpicture}
\caption{Conditional attack success versus patch scale.}
\label{fig:patchsize}
\end{figure}

Figure~\ref{fig:patchsize} sweeps the patch scale $a \in \{0.25, 0.5, 0.75, 1.0\}$. cASR generally rises with scale, though the increase is not monotonic. The ordering of vulnerability is broadly preserved across the sweep, meaning the adversarial patches are robust to scale and reflect vulnerability rather than the choice of that parameter.
 
\FloatBarrier
 
\subsection{IoU threshold ablation}
 
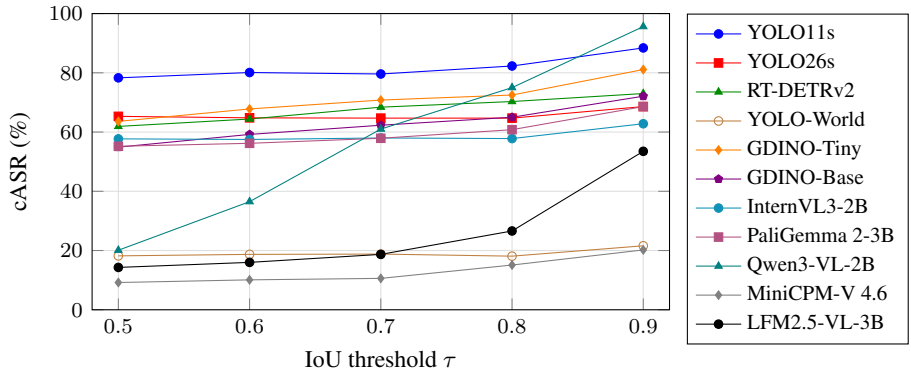
\begin{figure}[ht]
\centering
\begin{tikzpicture}
\begin{axis}[
    width=0.66\linewidth, height=5.5cm,
    xlabel={IoU threshold $\tau$},
    ylabel={cASR (\%)},
    xtick={0.5,0.6,0.7,0.8,0.9},
    xmin=0.48, xmax=0.92,
    ymin=0, ymax=100,
    legend style={at={(1.03,1)}, anchor=north west, font=\footnotesize},
    legend cell align={left},
    grid=major, grid style={gray!25},
    mark size=1.6pt,
    tick label style={font=\footnotesize},
    label style={font=\small},
]
\addplot[blue, mark=*]                 coordinates {(0.5,78.3)(0.6,80.1)(0.7,79.6)(0.8,82.3)(0.9,88.4)}; \addlegendentry{YOLO11s}
\addplot[red, mark=square*]            coordinates {(0.5,65.3)(0.6,64.8)(0.7,64.7)(0.8,64.7)(0.9,68.6)}; \addlegendentry{YOLO26s}
\addplot[green!55!black, mark=triangle*] coordinates {(0.5,61.9)(0.6,64.4)(0.7,68.4)(0.8,70.3)(0.9,73.0)}; \addlegendentry{RT-DETRv2}
\addplot[brown, mark=o]                coordinates {(0.5,18.2)(0.6,18.7)(0.7,18.8)(0.8,18.1)(0.9,21.6)}; \addlegendentry{YOLO-World}
\addplot[orange, mark=diamond*]        coordinates {(0.5,63.6)(0.6,67.8)(0.7,70.8)(0.8,72.5)(0.9,81.1)}; \addlegendentry{GDINO-Tiny}
\addplot[violet, mark=pentagon*]       coordinates {(0.5,54.9)(0.6,59.2)(0.7,62.3)(0.8,65.0)(0.9,72.1)}; \addlegendentry{GDINO-Base}
% VLM IoU-threshold sweeps
\addplot[cyan!70!black, mark=*]          coordinates {(0.5,57.7)(0.6,57.5)(0.7,58.0)(0.8,57.8)(0.9,62.8)}; \addlegendentry{InternVL3-2B}
\addplot[magenta!70!black, mark=square*] coordinates {(0.5,55.2)(0.6,56.2)(0.7,57.9)(0.8,60.8)(0.9,68.6)}; \addlegendentry{PaliGemma 2-3B}
\addplot[teal, mark=triangle*]           coordinates {(0.5,20.1)(0.6,36.5)(0.7,61.0)(0.8,75.0)(0.9,95.6)}; \addlegendentry{Qwen3-VL-2B}
\addplot[gray, mark=diamond*]            coordinates {(0.5,9.2)(0.6,10.1)(0.7,10.6)(0.8,15.1)(0.9,20.2)}; \addlegendentry{MiniCPM-V 4.6}
\addplot[black, mark=otimes*]            coordinates {(0.5,14.3)(0.6,16.0)(0.7,18.7)(0.8,26.6)(0.9,53.5)}; \addlegendentry{LFM2.5-VL-3B}
\end{axis}
\end{tikzpicture}
\caption{Conditional attack success versus IoU threshold.}
\label{fig:iou}
\end{figure}

Figure~\ref{fig:iou} sweeps the overlap threshold $\tau$ from 0.5 to 0.9. The size of the increase separates models by failure mode rather than family. Omission-dominated models show comparatively little change, because a box that is never emitted fails at every threshold. Models that always emit a box accumulate failures as the requirement tightens, with Qwen3-VL-2B rising from 20.1\% to 95.6\%.
 
\FloatBarrier
 
\subsection{VLM prompt ablation}
 
Table~\ref{tab:prompt} shows prompt sensitivity, re-evaluating each VLM patch under its training prompt and three paraphrases with the patch itself held fixed. The failure rate varies by prompt, so a single prompt should not be read as a model's robustness. However, the attacks mostly still generalize across different prompts, emphasizing deployment vulnerability. The failure partition remains stable: the gated formats remain omission-dominated under all four prompts, and Qwen3-VL-2B records exactly zero omissions under every prompt we test.
 
\begin{table}[ht]
\centering
\small
\setlength{\tabcolsep}{3pt}
\caption{Conditional attack success and failure partition across prompts.}
\label{tab:prompt}
\begin{tabular}{@{}>{\raggedright\arraybackslash}p{5.4cm}cccc@{}}
\toprule
Prompt & cASR & Omission & Substitution & Displacement \\
\midrule
\multicolumn{5}{@{}l}{\textbf{PaliGemma 2-3B} (gated)}\\
detect all people~(\textsc{trained}) & 55.2 & 53.4 & 0.0 & 1.9 \\
detect all pedestrians & 57.0 & 55.0 & 0.0 & 2.0 \\
detect all humans & 43.4 & 39.7 & 0.0 & 3.7 \\
detect all persons & 39.2 & 38.0 & 0.0 & 1.2 \\
\addlinespace
\multicolumn{5}{@{}l}{\textbf{InternVL3-2B} (gated)}\\
Please provide the bounding box coordinate of the region this sentence describes: \texttt{<ref>person</ref>}~(\textsc{trained}) & 57.7 & 56.7 & 0.0 & 1.0 \\
Please detect the person and return the bounding box. & 29.0 & 27.4 & 0.0 & 1.7 \\
Where is the person? Give the bounding box coordinates. & 45.9 & 39.3 & 0.0 & 6.7 \\
Locate the person and provide the bounding box coordinates. & 28.0 & 25.2 & 0.0 & 2.8 \\
\addlinespace
\multicolumn{5}{@{}l}{\textbf{MiniCPM-V 4.6} (free-text)}\\
Locate the person. Output ONLY the bounding box [x1, y1, x2, y2] in 0--1000 normalized coordinates, nothing else.~(\textsc{trained}) & 9.2 & 0.7 & 0.3 & 8.1 \\
Detect the person and output the bounding box [x1, y1, x2, y2] in 0--1000 coordinates. & 19.7 & 7.5 & 0.4 & 11.8 \\
Identify the person and give the bounding box [x1, y1, x2, y2] in 0--1000 coordinates. & 42.7 & 9.3 & 0.0 & 33.3 \\
Find the person and output the bounding box [x1, y1, x2, y2] in 0--1000 coordinates. & 30.1 & 7.3 & 1.2 & 21.6 \\
\addlinespace
\multicolumn{5}{@{}l}{\textbf{Qwen3-VL-2B} (free-text)}\\
Locate every person in the image and output the bounding box coordinates as JSON in the form \{``bbox\_2d'': [x1, y1, x2, y2], ``label'': ``person''\}.~(\textsc{trained}) & 20.1 & 0.0 & 1.5 & 18.7 \\
Locate the person and return \{``bbox\_2d'': [x1, y1, x2, y2], ``label'': ``person''\}. & 17.2 & 0.0 & 1.1 & 16.1 \\
Where is the person? Output \{``bbox\_2d'': [x1, y1, x2, y2], ``label'': ``person''\}. & 17.1 & 0.0 & 3.5 & 13.6 \\
Detect the person and output the bounding box as JSON \{``bbox\_2d'': [x1, y1, x2, y2], ``label'': ``person''\}. & 17.9 & 0.0 & 1.5 & 16.3 \\
\addlinespace
\multicolumn{5}{@{}l}{\textbf{LFM2.5-VL-3B} (free-text)}\\
Please provide the bounding box coordinate of the region this sentence describes: the person.~(\textsc{trained}) & 14.3 & 0.0 & 5.2 & 9.1 \\
Locate the person in the image and output the bounding box. & 1.6 & 0.0 & 0.0 & 1.6 \\
Detect the person and output the bounding box [x1, y1, x2, y2]. & 2.3 & 0.3 & 0.0 & 2.0 \\
Where is the person? Give the bounding box [x1, y1, x2, y2]. & 1.3 & 0.0 & 0.0 & 1.3 \\
\bottomrule
\end{tabular}
\end{table}
 
\FloatBarrier
 
\subsection{Detector confidence threshold ablation}
 
Table~\ref{tab:detsweep} varies the confidence threshold $\gamma$ at which each detector is read. As the confidence requirement increases, attack success increases for every model, suggesting that tasks that require more precision become more vulnerable.
 
\begin{table}[htbp]
\centering
\small
\setlength{\tabcolsep}{8pt}
\caption{Conditional attack success versus detector confidence threshold.}
\label{tab:detsweep}
\begin{tabular}{@{}lcccc@{}}
\toprule
Detector & \multicolumn{4}{c}{cASR (\%)} \\
\midrule
\multicolumn{5}{@{}l}{\textbf{Closed-set detectors}}\\
$\gamma$   & 0.1  & 0.25 & 0.5  & 0.7  \\
\cmidrule(l){2-5}
YOLO11s    & 67.2 & 73.1 & \textbf{78.3} & 83.8 \\
YOLO26s    & 40.7 & 56.7 & \textbf{65.3} & 67.9 \\
RT-DETRv2  & 34.7 & 49.8 & \textbf{61.9} & 68.6 \\
\addlinespace
\multicolumn{5}{@{}l}{\textbf{Open-vocabulary detectors}}\\
$\gamma$    & 0.1  & 0.2  & 0.3  & 0.4  \\
\cmidrule(l){2-5}
YOLO-World  & 11.1 & 16.8 & \textbf{18.2} & 23.1 \\
GDINO-Tiny  & 4.5  & 29.3 & \textbf{63.6} & 81.2 \\
GDINO-Base  & 3.6  & 22.5 & \textbf{54.9} & 77.4 \\
\bottomrule
\end{tabular}
\end{table}
 
\FloatBarrier

\section{Dataset construction}
\label{app:data}

\begin{figure}[ht]
\centering
\includegraphics[width=\linewidth]{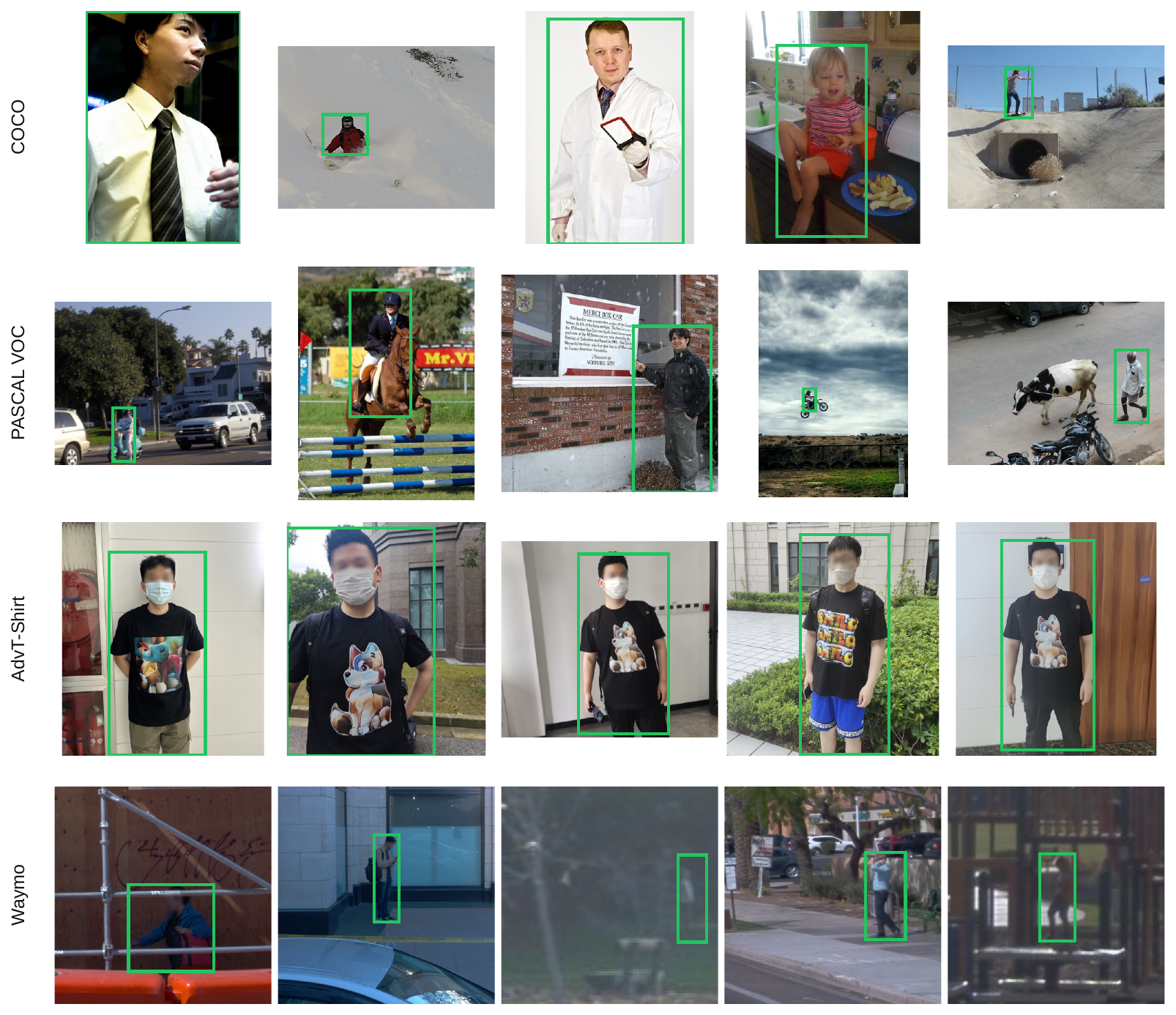}
\caption{Example images from each source dataset.}
\label{fig:datasets}
\end{figure}

 Table~\ref{tab:datasets} reports the filtering counts for each source dataset. From COCO we only take val2017, to prevent data leakage as some models are trained on train2017; from PASCAL VOC, the val sets of 2007 and 2012; from AdvT-Shirt, its 1-person folder; and from Waymo, the training split of v2 perception. Each dataset was first filtered to one-person images using its ground-truth bounding box labels. We then manually removed those containing extra unannotated people, and dropped Waymo's night-time images to ensure proper visibility. Figure~\ref{fig:datasets} shows example images from the final version of each dataset.
 
\begin{table}[ht]
\centering
\small
\caption{Per-dataset filtering counts.}
\label{tab:datasets}
\begin{tabular}{@{}lrrrr@{}}
\toprule
Source dataset & Original & One-person pool & Kept & Kept (\%) \\
\midrule
COCO val2017                        & 5{,}000  & 925   & 777 & 15.5 \\
PASCAL VOC 2007+2012 val       & 8{,}333 & 987   & 674 & 8.1  \\
AdvT-Shirt-1K, 1-person subset                   & 208      & 208   & 204 & 98.1 \\
Waymo v2 perception, training split & 4{,}662  & 1{,}014 & 999 & 21.4 \\
\bottomrule
\end{tabular}
\end{table}
 
\end{document}